\documentclass[letterpaper]{article}

\usepackage{natbib,alifeconf}  %% The order is important
\usepackage{url,hyperref,cleveref}
\usepackage{booktabs}

\usepackage{multicol}% http://ctan.org/pkg/multicols

\title{Evolved Developmental Artificial Neural Networks for Multitasking with Advanced Activity Dependence}

\author{
    Yintong Zhang$^{1}$ \and
    Jason A. Yoder$^{1}$
    \mbox{}\\
    $^1$Rose-Hulman Institute of Technology, USA \\
    yoder1@rose-hulman.edu
} % email of corresponding author

\begin{document}

\maketitle

\begin{abstract}
    % Abstract length should not exceed 250 words
    Recently, Cartesian Genetic Programming has been used to evolve developmental programs to guide the formation of artificial neural networks (ANNs). This approach has demonstrated success in enabling ANNs to perform multiple tasks while avoiding catastrophic forgetting. One unique aspect of this approach is the use of separate developmental programs evolved to regulate the development of separate soma and dendrite units. An opportunity afforded by this approach is the ability to incorporate Activity Dependence (AD) into the model such that environmental feedback can help to regulate the behavior of each type of unit. Previous work has shown a limited version of AD (influencing neural bias) to provide marginal improvements over non-AD ANNs. In this work, we present promising results from new extensions to AD. Specifically, we demonstrate a more significant improvement via AD on new neural parameters including health and position, as well as a combination of all of these along with bias. We report on the implications of this work and suggest several promising directions for future work.
\end{abstract}

\section{Introduction}
Artificial Neural Networks (ANNs) are central to the ongoing revolution in Artificial Intelligence today. Most models involving artificial neural networks (ANNs) use fixed-structure networks, and their connection weights are trained via gradient descent. Because the task information is mainly encoded in the connection weights of the artificial neural network (ANN), which can be overwritten by training with a new task, they are likely to undergo catastrophic forgetting when being used to solve more than one problem. 
On the other hand, biological neural networks show more flexibility, incorporating environmental feedback during development to solve many tasks at once.
Recently, more research has looked to biology for inspiration \citep{kudithipudi2022biological} for mechanisms that enable natural organisms to achieve flexible, adaptive intelligence desired by the field. 

According to the theory of Darwinian evolution, biological organisms start with simple structures and over time increase their complexity through development and evolution. Specifically, the process of development has allowed a wide range of organisms to grow from a single cell into multi-cellular organisms over an individual's lifespan. There are many factors that influence the developmental process including the internal and external environment, but also genetic factors. The genetic encoding that helps to regulate this process is still not well understood, but our knowledge and ability to model or imitate it continues to expand. What we do know is that in the natural world, many individuals undergo random mutations, and only sufficiently fit individuals are selected to survive and produce offspring. The phenotypes of these individuals are based on their genotypes and their environment, and their complex structures are produced through a sequence of developmental stages using the rules encoded by the genomes. In this paper, we look to expand upon an existing model \citep{miller_multi_2020} that attempts to simulate the evolution of a developmental process. Specifically, a developmental program to build a multitasking neural network is found using an evolutionary algorithm.

In this work, we reproduce comparable results to prior work while also simplifying the implementation. Most crucially, however, we conduct and report on results from experiments with an extended model of Activity Dependence (AD). AD is a mechanism that enhances the learning process of the neural network. AD allows the model to adjust its parameters based on feedback from tasks it performs. The key contribution of this work is the identification of the strong potential of previously unexplored targets of AD.

%Adding paragraph outlining the sections to follow
In the remainder of this paper, we will start by reviewing approaches to developmental neural networks and related fields. Following this, we will discuss in more depth the particular model we build upon including relevant differences in our models and important details for implementation. Next, we lay out the specific methodology for the experiments we conduct and the results of those experiments. Finally, we discuss the significance and implications of those results, concluding with a summary of the overall work and highlighting promising avenues for future research.

\section{Related Work }
In this section, we will provide a brief review of related and similarly motivated work. There have been a number of different approaches to exploring the space of possible neural network architectures along with synaptic weights such as Neural Architecture Search \citep{liu2021survey},  Hypernetworks \citep{chauhan2023brief}, and  neuroevolution\citep{stanley2019designing}. 

As a whole, neuroevolution has benefited from biology inspirations. The artificial neural network models in this field include varying forms of plasticity, although not necessarily as part of a developmental process \citep{mouret2014artificial}. These forms of plasticity have been utilized in a wide range of works \citep{soltoggio2018born}, however, many are focused primarily on learning mechanisms and less on development. Work focused on development has helped produce a taxonomy of artificial embryogeny \citep{stanley2003taxonomy} enabling classification of models along five key dimensions. 

By abstracting away the process of growth in development, Compositional Pattern Producing Networks (CPPNs) \citep{stanley2007compositional} manage to capture many of the desirable features of development, but in a highly efficient manner. CPPNs form the groundwork for the  HyperNEAT \citep{stanley2009hypercube} approach to neuroevolution, which is highly efficient and scalable. The underlying CPPN in HyperNEAT provides an indirect encoding of a neural network's structure and weights (as opposed to the direct encoding of the NEAT algorithm \citep{stanley2002evolving}). However, as suggested by \cite{hiesinger2021self}, the way that a developmental process unfolds over time (via local interactions) could be a crucial component enabling the complexity of biological networks such as the human brain.

A subset of the expanding collection of efforts to incorporate structural changes into the learning process of neural networks has been reviewed by \cite{maile2023structural}, who both provide connections to neuroscience, but also suggest a common neural operator framework to better relate different approaches. Despite these efforts, there have been relatively few attempts to allow neural networks which grow utilizing bottom-up approaches, with many of them presented in \cite{kowaliw2014growing} Several recent works have been inspired by the process of neurogenesis \citep{maile2022and, huang2023neurogenesis}. Another unique model is the HyperNCA \citep{najarro2022hypernca}, which trains a Neural Cellular Automata to guide a developmental process and acts as a Hypernetwork.

Aside from the work we directly build on in this paper \citep{miller_multi_2020}, the most closely related model is the Neural Developmental Program (NDP) \citep{najarro2023towards}, which also takes a bottom-up approach. In this approach a neural network guides the process of regulating the growth of another target neural network. We believe that our results imply strong promise for similar future extensions of the NDP model as well.

\section{Neural Network Model}
This section introduces the architecture of our model which builds upon an existing model \citep{miller_multi_2020}. Specifically, we focus on how the model encodes the developmental rules of ANNs, which allows them to be evolved by evolutionary algorithms. We also discuss the developmental process of a single ANN and how we added new features for Activity Dependence (AD). 

The model \citep{miller_multi_2020} we build upon focuses on ANN-generating programs evolved by evolutionary algorithms in the form of Cartesian Genetic Programming (CGP). The model takes inspiration from the same processes that have produced natural, biological brains, which include evolution, development, and learning. The model thus produces artificial feed-forward networks that are grown (development) according to rules from evolved programs (evolution) and adjusted by self-regulation via external feedback (learning). Past work demonstrated this model can solve multiple problems with a single network \citep{miller_multi_2020}. The model used Cartesian Genetic Programming (CGP) to transform encoded genomes into programs that grow a neural network. CGP builds a connection between genotypes encoded as strings and functioning computer programs by transforming one to the other, allowing the evolution of computer programs via the mutation of genotypes. Specifically, it uses numbers and strings to denote different kinds of computational operations and connections between inputs and outputs of operations. Any program encoded by CGP can be represented by a graph of operations, which can then be translated into string genotypes \citep{miller2008cartesian}. A key feature of Miller's system is the self-learning behavior realized by AD. AD adjusts internal neural variables based on external signals, such as task performance, similar to a reward signal in reinforcement learning. AD is one dimension of the model that has been previously explored, albeit in a limited fashion. More details about the original model and AD can be found in the original paper \citep{miller_multi_2020}.

\begin{figure}[tp] % CGPANN Figure
    \centering
    \includegraphics[width=3.5in]{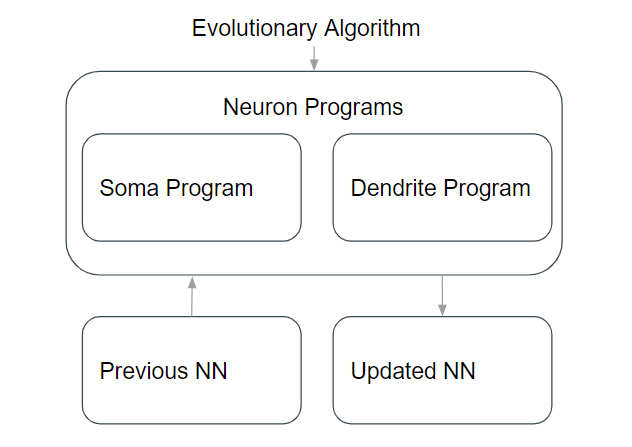}
    \caption{CGP ANN overview in which soma and dendrite refer, respectively, to nodes and connections in an artificial neural network.}
    \label{fig:cgp_ann}
\end{figure}

\begin{figure*}[tp] % neuron programs Figure
    \centering
    \includegraphics[width=7in]{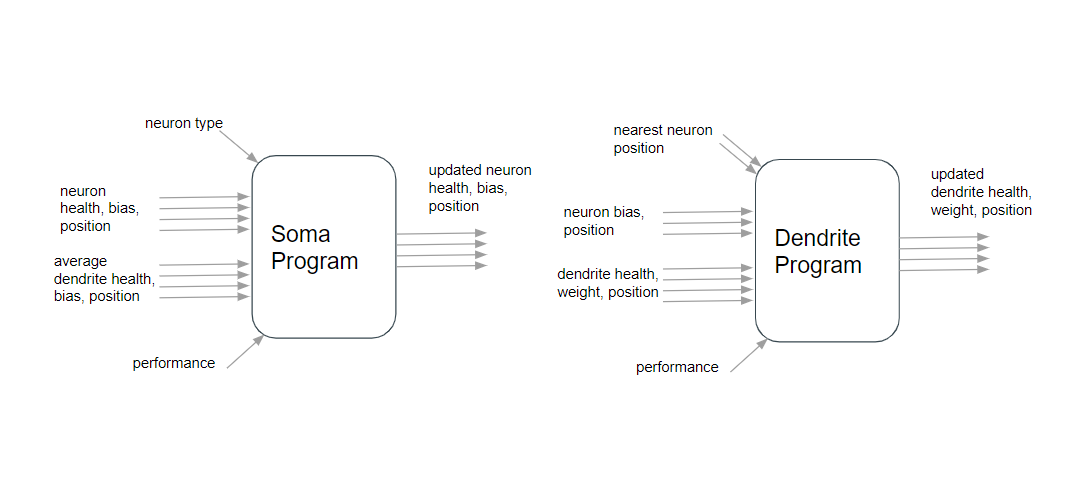}
    \caption{Inputs and Outputs for the Soma and Dendrite Program. The inputs come from parameters of nodes and connections in an ANN, and the outputs are fed back to the network as updated values for the same parameters.}
    \label{fig:neuron_programs}
\end{figure*}
Miller's model uses CGP to encode two programs as developmental rules for nodes (soma) and connections (dendrites). The populations, each involving two genotypes for both of the programs, are initialized randomly and they will be updated using the Evolutionary Algorithm. Each individual's initial neural network starts with a few input and output nodes in a 2-D space for the different tasks to solve. The output nodes have a few randomly initialized connections with their own positions in the space as well as other parameters including health and weight. In each developmental iteration, each node and connection are updated according to these programs and then the parameters of the whole network are generated for the next iteration. (see Figure~\ref{fig:cgp_ann}) Based on the output parameters, each of the nodes will have different properties such as x and y positions and health after each iteration. In this way, the structure of the neural network is updated due to the changing positions of nodes and some possible newly born nodes when a node has a high health value. After a predefined number of iterations, the final version of the network is regarded as a mature “brain” and sub neural networks are extracted for different problems by starting with output nodes for each problem and tracing back through dendrites and nodes. All dendrites of a node are connected to the node closest to the dendrite on the left. To avoid recurrent connections, dendrites with a position on the right of its mother node will be omitted. All extracted neural networks have nodes that accumulate the inputs from their dendrites with the respective weights and produce normalized outputs with tanh activation functions. Then, the networks can be used to solve multiple problems and performance can be measured.

In our implementation, however, sub-networks are not directly traced and extracted due to the computational cost. Instead, inputs of a single problem are fed into the corresponding input nodes in the original ANN, while other input nodes have no inputs. In this way, only part of the original ANN is invoked to solve one problem. Therefore, it maintains the same multitasking functionality as the original model. When tasks are performed, the original ANN is adjusted by AD regarding the signals present on sub-networks. As a result, the ANN learns to perform the tasks, and fitness values can be calculated. Finally, the individual with genotypes that develop the best-fit neural network is selected and reproduced for the next generation of evolution. Here, a one-elite reproduction strategy is used, which means the selected parent genotype remains in the next generation while other individuals are created by mutation of the elite genotype. The elitism provides a more robust evolutionary process by preserving the genotype from the previous generation if none of the random mutations improve performance.

 In the previous work, only the soma bias was adjusted by AD and it had marginal improvements compared to the base (non-AD) model \citep{miller_multi_2020}. In the scope of this work, we further explored the utility of AD in adjusting the parameters of soma. It includes the soma health, position (x and y), as well as the bias values. (see Figure~\ref{fig:neuron_programs}) We began by testing the AD bias to verify that our models can successfully reproduce the results from \citep{miller_multi_2020}. This likewise serves to allow for a fair comparison of the performance of the original model (AD bias only) to our alternative configurations. Among the soma parameters, The health of a soma can be updated in development cycles and is used to determine whether a soma should die or give birth to another soma in the next iteration. The relative positions between neurons determine their connection in the feed-forward network\citep{miller_multi_2020}. These parameters directly control the inputs and outputs of
 the somas as well as the overall structure of the network, therefore, AD adjustments on these parameters may produce more diverse results and visible effects on the performance compared to dendrite parameters (which can control connectivity, but cannot add, remove, or modify network nodes). 

\section{Methodology}
% 1. what are the tasks (brief justification)
% 2. what is the set up (population, selection, etc)
% 3. what are the key parameters used ( for reproducibility) 
In previous work \citep{miller_multi_2020}, the model was tested with AD neuron bias by solving four problems. Two of them were reinforcement learning (RL) problems and the other two were classification problems. In this work, we use one RL task and one classification task for testing. 

Our system was able to demonstrate solving both problems simultaneously.
We used the cartpole problem as our RL task and we used a bank authentication dataset \citep{Bank_Auth_Data} for the classification task. Although we did not use the exact same problems as the original work, we found that the combination of two kinds of problems produced comparable results for evaluating the ability of the new model.

For all experiments, an initially random population of size ten was used. At the end of the evolution loop, the individual with the best fitness was selected for producing the next generation. Specifically, we made nine mutated individuals based on the selected one, along with the original one unmutated (elite). All experiments were run for 100 generations. 

All other predefined parameters inside the model, if not specified here, follow the details specified in \citep{miller_noAD_2020} regarding the base model. For example, our experiments use the same number of development cycles and theta values for neuron creation/deletion as the original model used. Since the basic model architecture is the same, using the same parameters could prove the functionality and produce comparable results to the best effort. 

For our first experiment, we used the model without AD to solve a single problem to test its basic functionality. In solving the cartpole only, the model perfectly reaches a thousand steps and solves the problem within fifty generations (the classification was easier for the system to solve). Solving two problems simultaneously was more challenging for the neural network, as it only reached close to the maximum score of two thousand. However, it did make consistent progress toward the solution. We consider this sufficient to demonstrate our implementation has reproduced the functionality of the base model. With the base model performing the two tasks, we next tested the AD behavior on neuron bias, which showed similar improvements as those reported in \cite{miller_multi_2020}. This serves as a confirmation of the reproducibility of the previous findings about AD bias. After this, other neuron parameters were individually integrated and tested with the AD model. Finally, a combination of all of these parameters in the AD model was evaluated.

\section{Results}
\begin{figure}[tp] % CGPANN Figure
    \centering
    \includegraphics[width=3.5in]{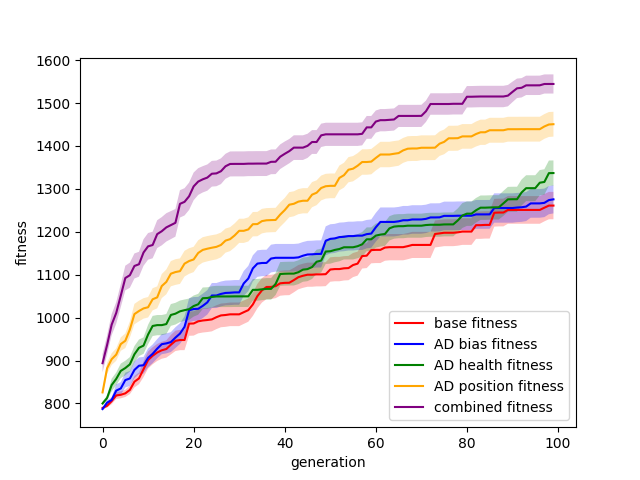}
    \caption{Comparing Performance for different Activity Dependence (AD) models. The base model includes no AD, and the combined model includes AD of all soma's parameters (bias, health, and position). Each experimental group was run 50 times and the lines show the mean of the best individuals from each run. The shaded area represents the standard error for each of the fitness lines.}
    \label{fig:exp}
\end{figure}

% 1. Introduce - what are the questions we seek to ask?
% 2. How do we achieve these results (define experiment (and any differences from the defined methodology)
%3. Objectively state what you can observe in the results - leave implications, meaning, and interpretation for the discussion section
For experimental results, we seek to compare the relative performances between experiments with different AD parameter configurations. In all cases, we run multiple experiments and report on the average score calculated on the best fitness of each run. Figure~\ref{fig:exp} provides summary plots of these results. 

Multiple experiments were conducted to compare the effects of AD behavior for different parameters present in the neural network. The experiments focused on soma parameters, which refer to the properties of nodes in a neural network: bias, health, and position (x and y). There are fifty evolutionary runs with each of the parameters modified by AD behavior, in which there are 100 generations within each run. Along with the fifty runs performed with the base model without AD, the effects of including AD on different parameters on the average performance can be directly compared to the base model. Finally, the AD behavior using all neuron parameters is evaluated to see the effects of a possible combination of AD, opening the door in the future 
 for fine-tuning or evolving AD behavior. The average performance from fifty runs during 100 generations is plotted in the graph (see Figure~\ref{fig:exp}). 

For the experiments in which AD influences neuron health or position, we see increased performances relative to the base model. While the plotted graphs have similar shapes, the performance scores for AD health and position are higher than the base model and AD bias at the end of 100 generations. Since they have similar starting points at around 800, the AD behaviors on a single parameter have been demonstrated to make the neural networks adapt to the problem more efficiently than without any AD. Finally, AD on all neuron parameters showed the greatest performance of all. The last result implies that AD behavior for all neuron parameters may have minimal interference with each other, allowing their improvements to accumulate and achieve a better performance score when solving the problems. 

\section{Discussion}
% 1. introductory paragraph - single sentence for each paragraph to follow
% 2. 

In the experiments, the results have shown promising improvements with different AD behaviors. Each of the experiments led to a noticeable increase in performance and a combination of all parameters could lead to larger improvements. However, there are several differences between the original model and the assumptions in our model, leading to uncertainties and limitations. 

As shown in Figure~\ref{fig:exp}, the performance curves for all experiments start from around 800 and approach a plateau, after which there is minimal improvement. The genotypes are generated and evolved randomly so that the limit of fitness is not the same every time, but the model is less likely to make notable progress after a hundred generations. A possible reason could be the experimental setup. In the experiments, the neural networks must deal with both the cartpole problem and a data classification problem. In the first one, the initial position and velocity of the cart are randomly initialized, leading to an unstable environment. Therefore, it is hard for the same model to get the perfect solution for cartpole because of the noise in the initial conditions. Furthermore, the challenge of this problem even affects the network’s ability to make correct decisions on the classification problem. 

In the experiments, one common best solution the model came up with was to solve the reinforcement learning problem while always predicting false on the second problem. This phenomenon implies that the model is likely to focus on the harder problem and make a simple guess on the easier one (i.e. always predict the same label for the classification problem). The available network complexity was all used for solving the harder cartpole problem and therefore achieving a higher overall accuracy even with the simple classifier for the first problem. In addition, the absence of some unspecified implementation details from the original work caused us to have to make certain assumptions which could also be limiting the quality of the final results.

\section{Conclusion and future work}
% try to summarize the areas in an introductory sentence, one sentence for each paragraph to follow
In this work, the implementation and experiments have focused on the improvement of learning efficiency in multitasking neural networks made by AD on neuron parameters. Although there are some uncertainties to be resolved in the experiments, we can safely conclude that AD with other neuron parameters leads to a greater performance increase compared to the previous work (Miller 2020). Moreover, combining AD behavior on different parameters can further improve the learning efficiency of this model. 

These results are promising and show notable improvements over the previous results. For related work, the results also increase the confidence of applying AD in other developmental models. For instance, a previous paper \citep{najarro2023towards} mentions the possibility of incorporating AD into the NDP model to enhance the model's learning ability. Therefore, the effects of different combinations of AD behavior could be explored with this model. On the other hand, a complete analysis of AD is not fully covered in this work. In addition to further analysis, we see great potential for future work in three directions: 1) exploring AD on other neural parameters, 2) optimizing combinations of parameters for AD, and 3) finding ways to increase the efficiency of the overall process.

% #1 expand the set of AD parameters
The first direction is to incorporate and evaluate the dendrite's parameters for AD behavior. By updating dendrite parameters using AD, the model can regulate the arrangement of connections between neurons, as well as the biases/weights of different dendrites. Given the results of AD on neuron parameters, this approach could further increase the learning efficiency of the model. Upon introducing this feature and evaluating the components, one could produce a baseline measurement of the individual utilities of the different AD neural parameters.

% #2 evolved AD
A second direction for the future is based on the observed benefits of combined AD behavior. Seeing potential synergy, or at least minimal interference, between different AD behaviors with neural parameters opens the possibility of AD being parameterized for specific neural features. One approach to searching such a parameter space is evolutionary algorithms (EA). Thus, one could use an EA to find an optimal AD combination to maximize learning capabilities. Therefore, an important direction for future work is introducing an evolvable AD behavior, which means using a genotype to represent the set of parameters to be changed for each neural network. 
 
% #3 consider other optimization approaches (list some potential ones, surrogate models come to mind but also maybe QD algorithms)
A final direction for future work is to focus on increasing the efficiency of the model. Although some steps in the algorithm have been simplified to speed up the execution, the model is still relatively slow due to the updating of a large brain network. Similarly, this model is inefficient because of the nature of evolutionary algorithms: every time the program is running with a large number of individuals only a small portion of them undergoes a good mutation. To address these problems, other applicable techniques, such as Quality Diversity \citep{pugh2016quality}, could be implemented to allow more systematic exploration of meaningfully different developmental behaviors.

\footnotesize
\bibliographystyle{apalike}
\bibliography{refs} % replace by the name of your .bib file

\end{document}